\documentclass[10pt,twocolumn,letterpaper]{article}

\usepackage[pagenumbers]{cvpr} 

\definecolor{cvprblue}{rgb}{0.21,0.49,0.74}
\usepackage[pagebackref,breaklinks,colorlinks,allcolors=cvprblue]{hyperref}

\title{Continuous Perception Matters: Diagnosing Temporal Integration Failures in Multimodal Models}

\author{Zeyu Wang$^{1}$,  Zhenzhen Weng$^{2}$\thanks{Work done while at Stanford University.},  Serena Yeung-Levy$^{1}$ \\
$^{1}$Stanford University, $^{2}$Waymo\\
\href{https://ai.stanford.edu/~zywang/projects/ContinuousPerceptionBenchmark/}{Website}
}

\begin{document}
\maketitle
\begin{abstract}

    Continuous perception, the ability to integrate visual observations over time in a continuous stream fashion, is essential for robust real-world understanding, yet remains largely untested in current multimodal models. We introduce CP-Bench, a minimal and fully controlled benchmark designed to isolate this capability using an extremely simple task: counting identical cubes in a synthetic scene while the camera moves and only reveals subsets of objects at any moment. Despite the simplicity of the setting, we find that state-of-the-art open-source and commercial models, including Qwen-3-VL, InternVL3, GPT-5, and Gemini-3-Pro, fail dramatically. A static-camera control variant confirms that the failure arises not from object recognition but from an inability to accumulate evidence across time. Further experiments show that neither higher sampling FPS, perception- or spatial-enhanced models, nor finetuning with additional videos leads to meaningful cross-temporal generalization. Our results reveal a fundamental limitation in modern multimodal architectures and training paradigms. CP-Bench provides a simple yet powerful diagnostic tool and establishes a clean testbed for developing models capable of genuine time-consistent visual reasoning.

\end{abstract}  
\section{Introduction}
\label{sec:intro}

\begin{figure}[t]
  \centering
  \includegraphics[width=0.95\linewidth]{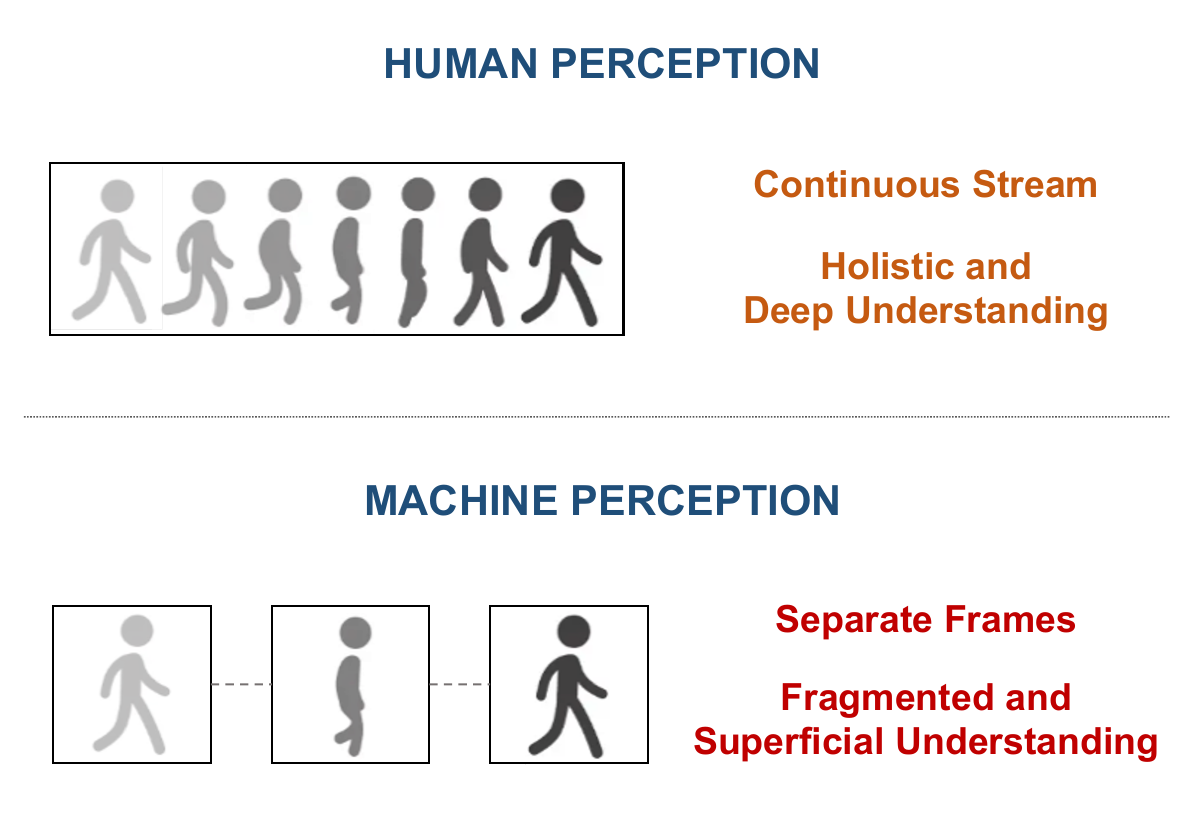}
  \caption{Human perception (top) operates on a continuous visual stream, which enables a holistic and deep understanding of spatio-temporal events. In contrast, dominant AI models (bottom) employ a separate frames paradigm, processing sparse, discrete snapshots. We argue this leads to a fragmented and superficial understanding that fundamentally fails to capture true spatio-temporal continuity.}
  \label{fig:pull}
\end{figure}

     Human vision operates as a continuous perceptual process: we experience the world through an uninterrupted stream of visual input, gradually accumulating evidence, establishing correspondence over time, and forming holistic, persistent representations of dynamic scenes (top of Figure~\ref{fig:pull})~\cite{deodato2024continuous,yi2008spatiotemporal}. This ability to integrate information across time is central to human visual intelligence, allowing us to track objects, reason about occlusions, and maintain stable scene understanding despite fragmentary views at any moment~\cite{deodato2022effect}.

    In contrast, contemporary vision-language and multimodal models~\cite{liu2024visual,zhang2023video,maaz2023video,lin2023video,team2023gemini,achiam2023gpt,chen2024internvl,wang2024internvideo2,wang2024qwen2,bai2025qwen2,zhu2025internvl3,comanici2025gemini,openai2024gpt5} process videos in a fundamentally different manner. Most existing architectures treat a video as a collection of separated frames, extracting features independently and aggregating them with shallow temporal modules (bottom of Figure~\ref{fig:pull})~\cite{liu2024visual,zhang2023video,maaz2023video,lin2023video,bai2025qwen2,chen2024internvl,wang2024internvideo2}. As a consequence, these systems tend to rely on texture-based shortcuts and momentary visual cues, resulting in fragmented and superficial video understanding. Despite rapid progress in video benchmarks and large-scale multimodal training, it remains unclear whether current models can perform the kind of continuous, stream-based perception that humans naturally employ.
    
    We argue that this capability has remained largely unexplored because existing video benchmarks~\cite{xiao2021next,mangalam2024egoschema,li2023mvbench,ning2023video,fu2025video,zhou2024mlvu,liu2024tempcompass,fang2024mmbench} rarely require continuous perception. Many benchmarks emphasize high-level semantics~\cite{patraucean2024perception}, action classification~\cite{kay2017kinetics,goyal2017something}, or summarization~\cite{kaushal2021good}, where success does not depend on integrating partial views over time. As a result, models can attain strong performance while entirely bypassing the deeper continuous temporal reasoning that characterizes human perception.

    To reveal this blind spot, we introduce the Continuous Perception Benchmark (CP-Bench), a deliberately minimalistic yet diagnostic evaluation of a model’s ability to integrate visual information over time. Each video consists of visually identical cubes placed on a plane, rendered with CLEVR-style~\cite{johnson2017clevr} physics and appearance, while a camera moves horizontally across the scene. At any moment, only a subset of cubes is visible, and the task is simply to count the total number of cubes. Because all cubes share identical shape, color, and texture, the task cannot be solved by single-frame appearance cues; instead, it inherently requires continuous perception for maintaining object permanence, aligning observations across frames, and constructing a global scene representation.
    
    Although CP-Bench is conceptually simple, our experiments reveal that it exposes a fundamental limitation of current multimodal systems. We evaluate a wide range of state-of-the-art open-source and proprietary models—including Qwen2.5-VL~\cite{bai2025qwen2}, Qwen3-VL~\cite{qwen3_vl2025}, InternVL3~\cite{zhu2025internvl3}, Gemini-3-Pro~\cite{comanici2025gemini}, and GPT-5~\cite{openai2024gpt5}, and find that none can reliably solve the benchmark. Even models that achieve nearly perfect accuracy in a static control setting, where all cubes are visible at once, fail dramatically under continuous observation. Increasing frame rate does not help, nor do models specifically finetuned for improved perception or spatial reasoning (e.g., PAPO~\cite{wang2025perception}, VST~\cite{yang2025visual}). While supervised finetuning on synthetic data can achieve high in-distribution accuracy, these models do not generalize across video durations or temporal configurations, indicating reliance on shortcuts rather than true temporal integration.
    
    These findings collectively suggest that current video-language architectures lack the mechanisms required for continuous, time-consistent perception. The failures persist across model families, training paradigms, temporal sampling densities, and specialized finetuning strategies. CP-Bench thus offers a uniquely simple yet powerful diagnostic tool for revealing this fundamental gap.
    
    In summary, our contributions are threefold:

    \begin{itemize}

        \item We identify continuous perception as a missing capability of modern video-language models, despite their strong performance on existing benchmarks.

        \item We introduce CP-Bench, an extremely simple yet revealing benchmark that isolates continuous perception from confounding cues.

        \item We provide extensive empirical evidence, across model architectures, closed- and open-source systems, FPS settings, and finetuning regimes, demonstrating that current models fail to perform continuous perception and instead rely on brittle shortcuts.
    \end{itemize}

    Our results point toward the need for new architectures and training paradigms explicitly designed to model continuous spatiotemporal processes. We hope that CP-Bench serves as a foundation for future research toward truly human-like video understanding.

\section{Related Work}
\label{sec:related}

\subsection{Multimodal Foundational Models}

    Recent years have witnessed significant progress in multimodal foundation models that integrate visual and textual modalities to achieve advanced understanding and reasoning capabilities~\cite{liu2024visual,zhang2023video,maaz2023video,lin2023video,team2023gemini,achiam2023gpt,chen2024internvl,wang2024internvideo2,wang2024qwen2,bai2025qwen2,zhu2025internvl3,comanici2025gemini,openai2024gpt5}. Open-source efforts such as the Intern-VL series (including InternVL~\cite{chen2024internvl}, InternVL2~\cite{wang2024internvideo2}, 2.5~\cite{chen2024expanding}, and 3~\cite{zhu2025internvl3}) and the QwenVL series (Qwen-VL~\cite{bai2023qwenvl}, Qwen-2-VL~\cite{wang2024qwen2}, Qwen-2.5-VL~\cite{bai2025qwen2}, and Qwen-3-VL~\cite{qwen3_vl2025}) have demonstrated steady improvements in scaling, architecture design, and multimodal comprehension. Concurrently, closed-source models like OpenAI’s GPT-5~\cite{openai2024gpt5} and Google’s Gemini-3~\cite{comanici2025gemini} have pushed the frontier of multimodal vision-language reasoning and generation, achieving impressive performance on a range of complex benchmarks. Despite these successes, our work highlights a critical gap: these models fundamentally lack the human-like ability for continuous perception, i.e., the temporal integration of partial visual observations into a coherent and holistic understanding over time. This shortcoming limits their true video understanding and spatial intelligence.

\subsection{Video Benchmarks}

    The evaluation of video understanding has been facilitated by a variety of recently proposed benchmarks~\cite{xiao2021next,mangalam2024egoschema,li2023mvbench,ning2023video,fu2025video,zhou2024mlvu,liu2024tempcompass,fang2024mmbench} that capture diverse temporal and multimodal challenges. Widely used datasets such as MMBench-Video~\cite{fang2024mmbench},  MVBench~\cite{li2023mvbench}, VideoMME~\cite{fu2025video}, and TempCompass~\cite{liu2024tempcompass} provide comprehensive platforms for assessing video representation learning, complex temporal reasoning, and multimodal fusion. These benchmarks are invaluable for pushing model performance on high-level semantic tasks. However, their complexity can often obscure fundamental failures. In contrast to these increasingly complex benchmarks, the Continuous Perception Benchmark (CP-Bench) proposed in this paper is intentionally minimalist. We do not aim to test high-level semantics or long-form reasoning. Instead, the CP-Bench is designed as a targeted diagnostic to probe the most basic foundation of visual understanding: the ability to maintain spatio-temporal correspondence across a continuous visual stream. By stripping away semantic complexity and visual variety, the CP-Bench isolates this single, critical capability, revealing a failure mode that is overlooked by more complex evaluations.

\section{Continuous Perception Benchmark}
\label{sec:benchmark}

\begin{figure*}[t]
  \centering
  \includegraphics[width=0.95\linewidth]{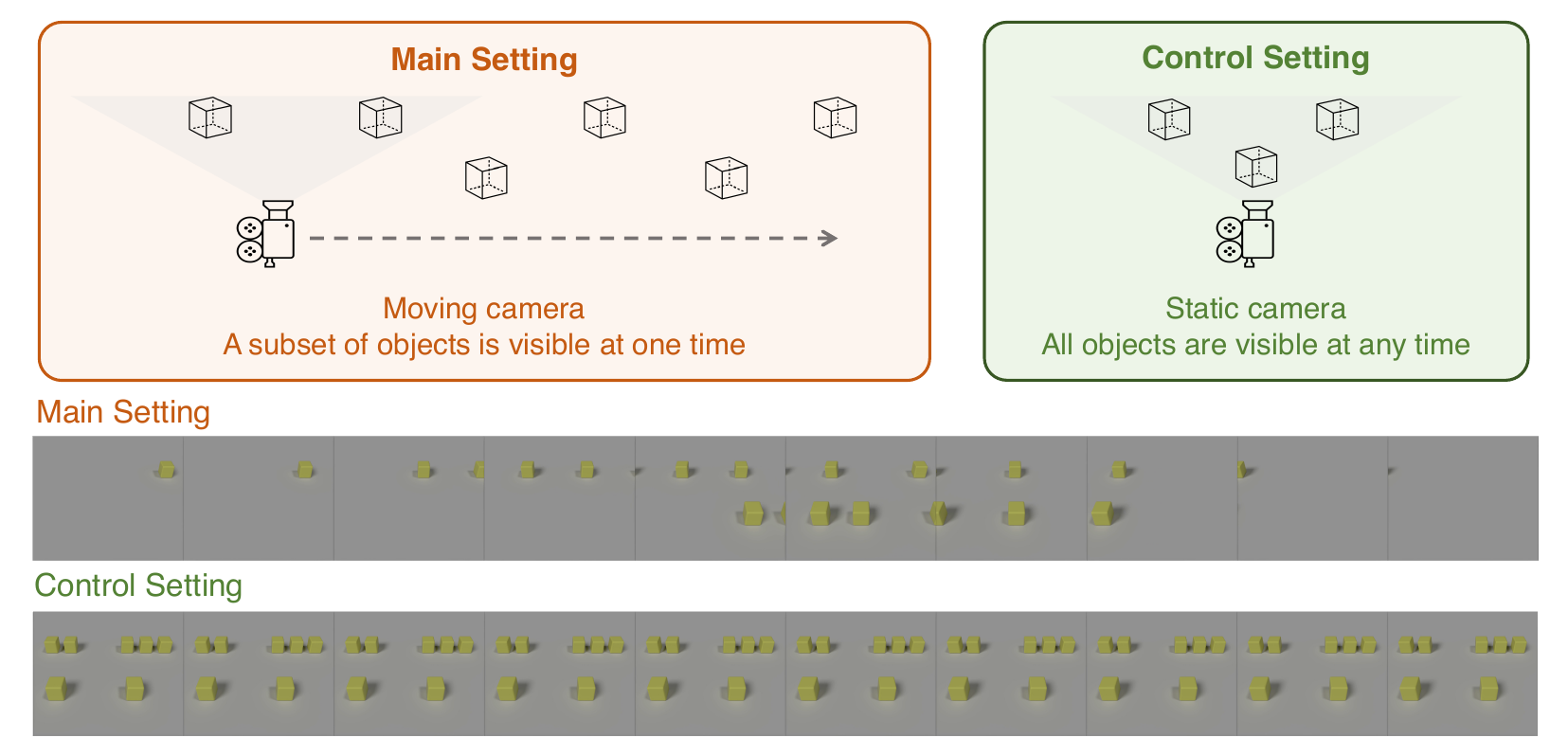}
  \caption{Illustration of the proposed Continuous Perception Benchmark (CP-Bench). In the Main Setting (top-left), a camera performs a continuous horizontal pan, with only a subset of visually identical cubes visible at any given moment. This necessitates continuous spatio-temporal correspondence for accurate counting. In the Control Setting (top-right), the camera is static, and all cubes are visible simultaneously, allowing for static-frame counting. The bottom panel provides visual examples of sampled frames from typical videos.}
  \label{fig:benchmark}
\end{figure*}

To investigate whether contemporary video models possess the ability to perform continuous, stream-based perception, we introduce the Continuous Perception Benchmark (CP-Bench). Unlike existing video benchmarks that emphasize high-level semantics, complex scenes, or domain-specific priors~\cite{xiao2021next,mangalam2024egoschema,li2023mvbench,ning2023video,fu2025video,zhou2024mlvu,liu2024tempcompass,fang2024mmbench}, CP-Bench isolates a single fundamental capability: constructing a coherent, persistent understanding of a scene from temporally continuous observations. Our goal is to provide an extremely minimalistic yet diagnostic testbed that reveals whether a model can move beyond fragmented frame-level processing toward holistic, temporally integrated perception.

\paragraph{Task Design.} CP-Bench is built on the CLEVR~\cite{johnson2017clevr} dataset, using Blender~\cite{blender} to render synthetic scenes. Each scene consists of a static planar surface populated by a set number of cubes. All cubes share identical appearance, i.e are identical in shape, color, size, material, and texture. This is a critical design choice made to eliminate texture-based shortcuts and prevent recognition based on superficial visual cues~\cite{geirhos2020shortcut}. As shown in top left of Figure~\ref{fig:benchmark}, the virtual camera performs a single, smooth horizontal panning motion across the scene over a duration of 10 seconds. The camera's field of view is constrained, ensuring that only a subset of the total cubes is visible at any given moment. All rendering parameters other than cube positions are kept constant across scenes. The task is to answer the question: ``How many cubes are present in the scene?''. To correctly infer the total number of cubes present on the plane, a model must maintain cross-frame correspondence, integrate partial observations over time, thereby engaging genuine continuous perception.

\paragraph{Statistics.} The benchmark includes five object-count configurations: 3, 4, 5, 6, and 7 cubes. We generated 20 unique instances for each count, resulting in a balanced diagnostic set of 100 total video instances.

\paragraph{Rationale.} CP-Bench is intentionally designed to be as simple as possible in appearance yet as revealing as possible in cognitive demand. By stripping away high-level semantics, object categories, motion complexity, and texture discriminability, the benchmark enforces reliance on temporal continuity rather than pattern matching. Unlike many modern video benchmarks that contain rich but confounded signals, CP-Bench isolates the core challenge that human vision excels at but current models struggle with: forming stable, persistent scene representations across time. CP-Bench is trivially easy for a human but, as shown later in the experiment section, exceptionally difficult for existing models. Its minimalism ensures that failure is not due to a lack of semantic knowledge or action-composition understanding, but rather a fundamental failure in the mechanism of perception itself. 
\section{Experiments}
\label{sec:experiments}

\subsection{Experiment Setup}

    To assess whether current state-of-the-art multimodal models possess continuous perception capabilities, we evaluate a broad set of representative systems from both the open-source and proprietary ecosystems.
    
    \paragraph{Open-Source Models.} We include three generations of the Qwen vision-language family, Qwen2-VL-7B~\cite{wang2024qwen2}, Qwen2.5-VL-7B~\cite{bai2025qwen2}, and Qwen3-VL-8B~\cite{qwen3_vl2025}, as well as the InternVL series comprising InternVL2-8B~\cite{wang2024internvideo2}, InternVL2.5-8B~\cite{chen2024expanding}, and InternVL3-8B~\cite{zhu2025internvl3}. These models cover the latest publicly available video-capable architectures and reflect the rapid evolution of open-source multimodal modeling.
    
    \paragraph{Proprietary Models.} We additionally evaluate leading commercial systems: GPT-5-mini and GPT-5~\cite{openai2024gpt5} from OpenAI, and Gemini-2.5-Flash, Gemini-2.5-Pro and Gemini-3-Pro~\cite{comanici2025gemini} from Google. These represent the most advanced closed-source multimodal models available at the time of our study, providing a comprehensive view of the current upper bound of video understanding performance.
    
    \paragraph{Evaluation Protocol.}  For all models (except Gemini which handles videos internally), we adopt a standardized 1 fps sampling rate. Each model receives the same set of frames extracted uniformly from the 10-second videos in CP-Bench. To facilitate reliable and unambiguous parsing of model outputs, we pose the counting task using a fixed prompt: ``How many cubes are present in the scene? Provide your response as a single numerical value.'' The predicted integer is compared directly against the ground-truth object count. We report accuracy as the primary evaluation metric, computed as the percentage of video instances for which the model outputs the correct total number of cubes.
    
    \paragraph{Control Experiment.} To further eliminate potential confounding factors and verify that model failures on CP-Bench are not attributable to low-level perceptual limitations, we design a control test set that preserves the exact visual appearance of the main benchmark while removing the need for continuous perception.
    
    In this control condition, the camera remains static, and all cubes are simultaneously visible in every frame, as shown in top right of Figure~\ref{fig:benchmark}. Apart from the absence of camera motion, all rendering settings remain identical to those used in the main continuous perception benchmark. The control set contains the same five ground-truth object counts (3–7), with 20 instances per count, for a total of 100 test videos.
    
    This control experiment is designed to eliminate key confounding variables by verifying that models can both successfully recognize the synthetic, textured cubes and perform basic counting of 3 to 7 objects in an unobstructed scene. If models succeed on this static control set but fail on the panning main setting, it provides strong evidence that the failure mode is rooted specifically in the lack of spatio-temporal continuity processing, not in fundamental recognition or counting errors.

\begin{table*}[t]
      \centering
      \begin{tabular}{l|ccccc|c|c|c}
        \toprule
        Model & GT=3 & GT=4 & GT=5 & GT=6 & GT=7 & Overall & Control  & $\Delta$\\ 
        \midrule

InternVL2-8B	&  30.0	&  0.0	&  0.0	&  0.0	&  0.0	&  6.0	&  41.0	& $\downarrow$ 35.0  \\
InternVL2.5-8B	&  25.0	&  0.0	&  0.0	&  0.0	&  0.0	&  5.0	&  58.0	& $\downarrow$ 53.0  \\
InternVL3-8B	&  30.0	&  0.0	&  0.0	&  0.0	&  0.0	&  6.0	&  79.0	& $\downarrow$ 73.0  \\
Qwen2VL-7B 	&  30.0	&  60.0	&  65.0	&  20.0	&  15.0	&  38.0	&  58.0	& $\downarrow$ 20.0  \\
Qwen2.5VL-7B	&  25.0	&  80.0	&  45.0	&  35.0	&  15.0	&  40.0	&  86.0	& $\downarrow$ 46.0  \\
Qwen3VL-8B	    &  35.0	&  65.0	&  20.0	&  20.0	&  10.0	&  30.0	&  82.0	& $\downarrow$ 52.0  \\
\midrule
GPT-5-Mini  	&  60.0	&  20.0	&  0.0	&  15.0	&  0.0	&  19.0	&  99.0	& $\downarrow$ 80.0  \\
GPT-5       	&  25.0	&  5.0	&  0.0	&  5.0	&  5.0	&  8.0	&  97.0	& $\downarrow$ 89.0  \\
Gemini-2.5-Flash	&  50.0	&  15.0	&  15.0	&  15.0	&  15.0	&  22.0	&  100.0	& $\downarrow$ 78.0  \\
Gemini-2.5-Pro	&  40.0	&  35.0	&  25.0	&  30.0	&  15.0	&  29.0	&  100.0	& $\downarrow$ 71.0  \\
Gemini-3-Pro &  80.0	&  55.0	&  55.0	&  40.0	&  25.0	&  51.0	&  100.0	&  $\downarrow$ 49.0  \\
        
        \bottomrule
      \end{tabular}
      \caption{Main experimental results on the Continuous Perception Benchmark (CP-Bench) and the Static Control set. All models are evaluated at 1 FPS. We report accuracy (\%) broken down by ground truth (GT) count, the ``Overall'' CP-Bench accuracy, the ``Control'' set accuracy, and the performance drop ($\Delta$). The results highlight a universal failure on the CP-Bench despite strong performance on the control task.}
      \label{tab:main_res}
\end{table*}

\subsection{Experiment Results}

\subsubsection{Main Experiment}

    Table~\ref{tab:main_res} presents the performance of all evaluated models on CP-Bench, along with their corresponding accuracy on the control test set and the performance gap ($\Delta$). Across all models, open-source and proprietary, we observe a striking and consistent pattern: \textbf{none of the models are able to reliably infer the total number of cubes when continuous temporal integration is required.}

    Open-source models from the InternVL family exhibit severe failure under the continuous setting, with overall accuracies between 5–6\%, despite achieving substantially higher performance on the control set (41–79\%). Qwen models perform moderately better, yet still fall far short of human-level expectations. Even the strongest open-source variant, Qwen2.5-VL-7B, reaches only 40\% accuracy, far below what would be required for correct temporal aggregation.

    Proprietary models, including GPT-5, GPT-5-Mini, Gemini-2.5-Flash, Gemini-2.5-Pro and Gemini-3-Pro, similarly fail to solve the benchmark. Despite attaining near-perfect accuracy on the control set (97–100\%), their performance on the main benchmark remains surprisingly low, with overall accuracies ranging from 8\% to 51\%. Notably, the gap between the main benchmark and the control setting is dramatic: 80–89 percentage points for GPT models and 49–78 points for Gemini models.

    These results collectively demonstrate that state-of-the-art multimodal systems, despite their impressive abilities on existing video benchmarks, lack the capacity for continuous perception. The dramatic accuracy drop compared to the control condition suggests that failure is not due to object recognition ability or difficulty in counting identical cubes. Rather, the models struggle to maintain object permanence, form temporally stable correspondences, and integrate sequential observations into a coherent global representation of the scene.

\begin{figure*}[t]
  \centering
  \includegraphics[width=0.95\linewidth]{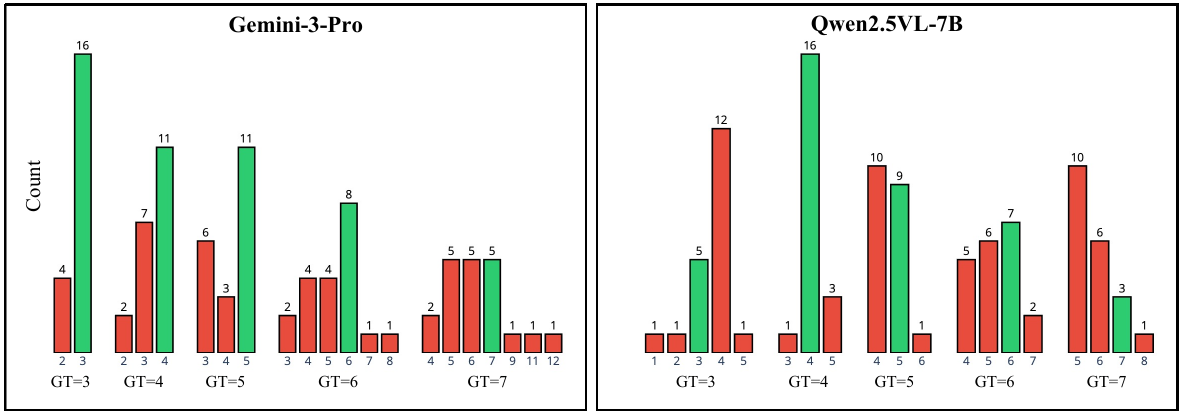}
  \caption{Prediction distribution for Gemini-3-Pro (left) and Qwen2.5VL-7B (right) on the Continuous Perception Benchmark (CP-Bench). Green bars indicate correct predictions, while red bars represent incorrect predictions. The X-axis shows the model's predicted count, grouped by the ground truth (GT) count for each set of instances.}
  \label{fig:bar}
\end{figure*}

    Figure~\ref{fig:bar} provides a detailed breakdown of the prediction distributions for Gemini-3-Pro and Qwen2.5VL-7B on the CP-Bench. The green bars indicate correct predictions, while red bars show incorrect ones.

    Gemini-3-Pro’s predictions exhibit a strong tendency toward undercounting. For ground-truth (GT) counts of 3, 4, and 5, the model produces only undercount errors. At GT = 6, it undercounts in 10 cases (50\%) and overcounts in 2 cases (10\%). At GT = 7, the model undercounts in 12 cases (60\%) and overcounts in 3 cases (15\%). These results indicate that the model often establishes incorrect correspondences, interpreting newly appearing objects as previously observed ones, thereby contributing to systematic underestimation. 
    
    Qwen2.5VL-8B exhibits a different prediction pattern. It achieves exceptionally high accuracy for GT=4 (16 correct instances, 80\%) but performs poorly on other counts. Notably, the model tends to overcount when GT=3, while consistently undercounting for GT=5, 6, and 7. This behavior suggests a strong bias towards the number `4', which may stem from a language prior.

\subsubsection{Effect of Temporal Sampling Rate}

    The main experiment evaluates all models under a uniform 1 fps frame sampling rate. To examine whether denser temporal sampling improves performance, potentially enabling models to approximate continuous perception by observing more frames, we conduct an additional experiment varying the input frame rate from 1 to 5 fps. Table~\ref{tab:fps_inference} summarizes the overall accuracy of each open-source model across different FPS settings.

    Results show that increasing the frame rate does not meaningfully improve performance. For the InternVL series, raising FPS either has no effect or further degrades accuracy. All models remains single-digit performance regardless of sampling density. The Qwen models show similarly behavior. Qwen2-VL-7B maintains moderate performance around 38–41\% for FPS 1–3 but drops sharply at higher frame rates. Qwen2.5-VL-7B fluctuates without showing any monotonic changes, and Qwen3-VL-8B degrades dramatically as FPS increases. Across all models, no configuration yields a meaningful accuracy gain relative to the 1 fps baseline.
    
    These results provide further evidence that the failure of existing multimodal models is not due to insufficient temporal sampling, but rather due to the lack of mechanisms for long-range temporal integration and object permanence. Simply supplying more frames does not help if the model processes them in a fragmented, frame-wise manner without forming a persistent global scene representation. This supports our central hypothesis:\textbf{ contemporary video models do not perform continuous perception, and increasing input density alone cannot compensate for this architectural and algorithmic limitation.}

\begin{table}[t]
      \centering
      \small
      \begin{tabular}{lccccc}
        \toprule
        Model & FPS-1 & FPS-2 & FPS-3 & FPS-4 & FPS-5 \\ 
        \midrule 
        InternVL2-8B & 6.0 & 1.0 & 1.0 & 1.0 & 1.0 \\
        InternVL2.5-8B & 5.0 & 5.0 & 5.0 & 5.0 & 5.0 \\
        InternVL3-8B & 6.0 & 6.0 & 8.0 & 9.0 & 9.0 \\
        \midrule
        Qwen2VL-7B & 38.0 & 41.0 & 41.0 & 34.0 & 22.0 \\
        Qwen2.5VL-7B & 40.0 & 38.0 & 23.0 & 27.0 & 33.0 \\
        Qwen3VL-8B & 30.0 & 12.0 & 11.0 & 7.0 & 8.0 \\
        \bottomrule
      \end{tabular}
      \caption{Overall Accuracy (\%) on the Continuous Perception Benchmark (CP-Bench) when evaluating models at different inference frame sampling rates (FPS). The results demonstrate that increasing the temporal density does not lead to a consistent improvement and, in several cases, degrades performance.}
      \label{tab:fps_inference}
\end{table}

\subsubsection{Effect of Perception-Enhanced Models}

    Recent work has proposed specialized training strategies to improve multimodal reasoning, perceptual robustness, and visuospatial understanding in vision-language models~\cite{wang2025perception,yang2025visual}. To further investigate whether stronger perceptual or visuospatial training can improve performance on CP-Bench, we evaluate two families of models built on top of Qwen2.5-VL: PAPO~\cite{wang2025perception} and VST~\cite{yang2025visual}. Both methods explicitly aim to enhance perception or spatial reasoning, and thus provide a natural testbed for examining whether such improvements translate into continuous perception abilities.

    PAPO (Perception-Aware Policy Optimization)~\cite{wang2025perception} introduces perception-oriented objectives into multimodal RL to reduce perceptual errors and enhance visually grounded reasoning. We test two variants, PAPO-G-7B and PAPO-D-7B. VST (Visual Spatial Tuning)~\cite{yang2025visual} aims to improve visuospatial understanding through a large-scale spatial perception dataset and a progressive SFT→RL training pipeline. We evaluate VST-SFT-7B and the full VST-RL-7B model.

    The results are presented in Table~\ref{tab:other_model}. All four model variants perform well on the static control task, achieving accuracies between 80.0\% and 93.0\%. Three of them, PAPO-G, PAPO-D, and VST-SFT further improve over their base model. This confirms their robust capability for static object recognition and counting.

    However, on the proposed CP-Bench, all enhanced models fail. More significantly, they all perform considerably worse than their own base model, Qwen2.5VL-7B (which scored 40.0\%). The PAPO variants, despite being optimized for perception-awareness, collapse to 17.0\% accuracy. The VST models, even after extensive spatial tuning (VST-SFT) and subsequent reinforcement learning (VST-RL), only reach 21.0\% and 30.0\%, respectively.

    This result demonstrates that the bottleneck is not a simple lack of ``perception'' or ``spatial skill'' that can be patched with current finetuning techniques. Even models explicitly trained to be better at perception and spatial reasoning are still fundamentally constrained by their discrete-frame-processing architecture. They have not learned the mechanism of continuous perception. This suggests that the ``spatial reasoning'' these models (like VST) have learned is likely still bound to fragmented-frame analysis, failing to generalize to our task which demands integrating a continuous spatio-temporal stream.

\begin{table}[t]
      \centering
      \begin{tabular}{lcc}
        \toprule
        Model & Overall & Control \\ 
        \midrule 
        Qwen2.5VL-7B & 40.0 & 86.0 \\
        \midrule
        PAPO-G-7B & 17.0 & 89.0 \\
        PAPO-D-7B & 17.0 & 93.0 \\
        VST-SFT-7B & 21.0 & 91.0 \\
        VST-RL-7B & 30.0 & 80.0 \\
        \bottomrule
      \end{tabular}
      \caption{Performance comparison of enhanced models (PAPO and VST) against their base model, Qwen2.5VL-7B. We report Overall Accuracy (\%) on the CP-Bench and the Control set. Despite being fine-tuned for ``perception'' (PAPO) and ``spatial reasoning'' (VST), the models show a performance drop on our continuous task, indicating their enhancements do not address the core challenge of continuous perception.}
      \label{tab:other_model}
\end{table}

\subsubsection{Finetuning on Synthetic Training Data}

    To further probe whether current models can learn continuous perception when provided with supervised data, we generate a synthetic training set that mirrors the structure of CP-Bench. Using the same rendering pipeline, we create 1,000 training videos: 200 instances for each ground-truth count from 3 to 7. All scenes strictly avoid overlap with test-set configurations, ensuring a clean separation between training and evaluation. We then finetune Qwen2.5-VL-7B for 5 epochs using a 3e-5 learning rate, cosine decay scheduling, with warm-up on first 5\% training iterations. Following common practice, we keep the visual encoder frozen during finetuning.

    The finetuned model achieves 89\% accuracy on the CP-Bench test set, which is significantly higher than the 40\% accuracy of the original model. At first glance, this might suggest that the model has successfully acquired continuous perception capabilities. However, the extreme simplicity and uniformity of our setting raise the possibility that the model may have discovered shortcut strategies instead of learning to integrate information across time. The task’s visual environment is highly constrained, and under such constraints, it is plausible that the model overfits to statistical regularities rather than genuinely reasoning over temporal continuity.

\begin{figure}[t]
  \centering
  \includegraphics[width=0.95\linewidth]{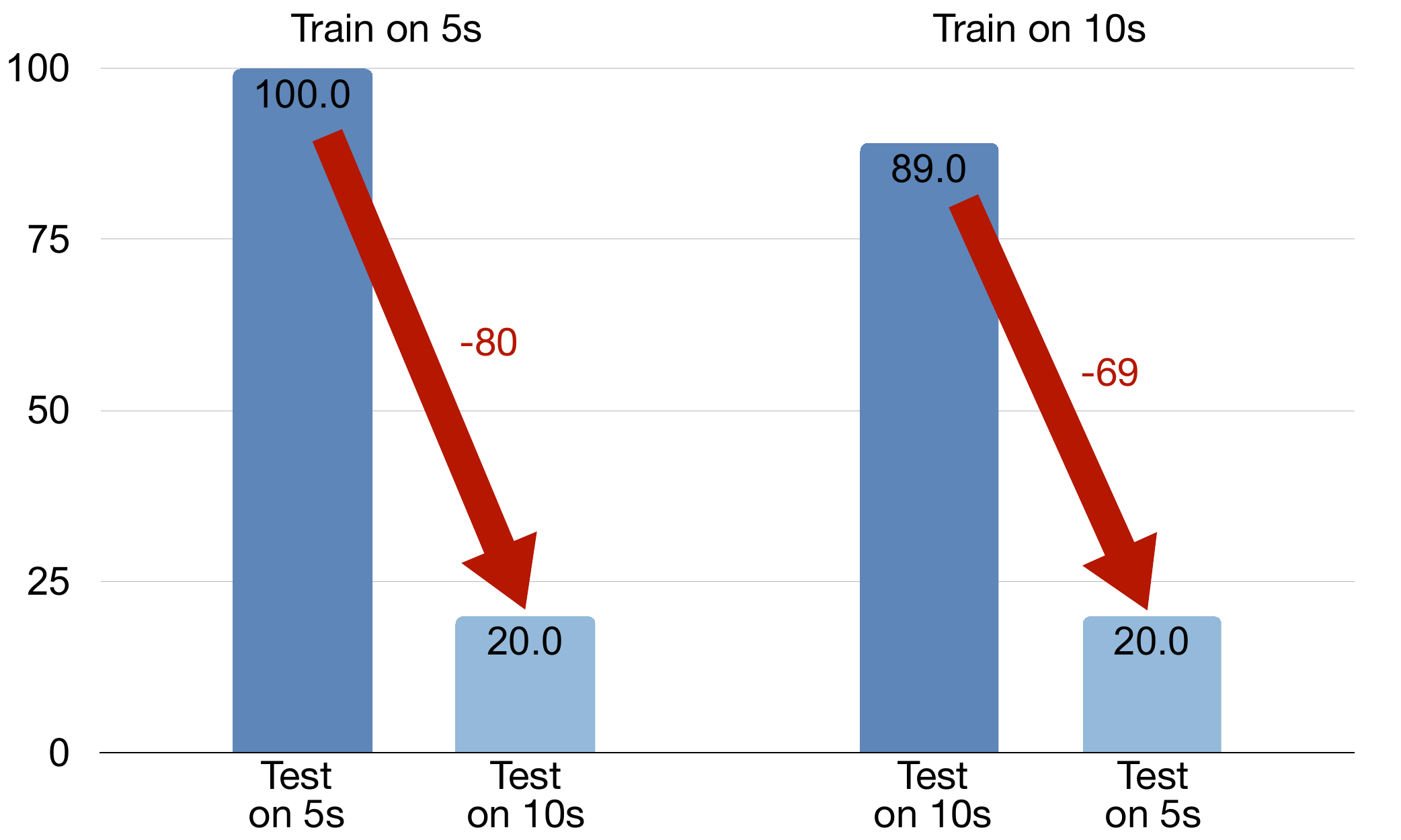}
  \caption{Fine-tuning generalization experiments. Models trained on 5s videos achieve perfect accuracy on the 5s test set but fail to generalize to 10s videos (left, 80-point drop). Conversely, models trained on 10s videos perform well on the 10s test set but fail to generalize to 5s videos (right, 69-point drop). This demonstrates that fine-tuning leads to overfitting to specific video dynamics (e.g., duration, camera speed) rather than learning the generalizable skill of continuous perception.}
  \label{fig:generalization}
\end{figure}

    To distinguish true continuous perception from shortcut-based learning, we modify the training set by shortening video duration from 10 seconds to 5 seconds, effectively halving the camera’s traversal distance. The test set remains unchanged (10 seconds). If the model had learned a temporally generalizable perception strategy, then reducing training video length should not substantially impact test performance. However, under this setting, the finetuned model achieves only 20\% accuracy on the 10-second test set, a drastic drop from the earlier 89\%, as shown in the right part of Figure~\ref{fig:generalization}.

    We also perform the reverse experiment: training on 10-second videos and testing on 5-second videos. Again, the model achieves only 20\% accuracy while training on 5-second achieves perfect accuracy, as shown in the left part of Figure~\ref{fig:generalization}.

    Across both experiments, models show poor cross-duration generalization, revealing that simple supervised finetuning, even when using data tailored for the task, does not yield continuous perception. The model does not acquire the ability to maintain a persistent scene representation over arbitrary temporal horizons; rather, it appears to rely on duration-specific patterns present in the training set. This result not only reinforces our core thesis that current architectures and training paradigms fundamentally lack mechanisms for robust continuous perception, but also highlights that this train-test-generalization setup provides a simple testbed for distinguishing genuine continuous perception from superficial overfitting, facilitating development of models explicitly designed to maintain temporally cohesive representations.

\subsubsection{Training with Higher Frame Rates}

    In previous experiments, models failed to generalize across different video durations, suggesting that simple finetuning does not induce continuous perception. One possible hypothesis is that the limited temporal density of training videos (e.g., 1 fps) might prevent models from learning stable temporal correspondences. If so, increasing the training frame rate, thereby providing more densely sampled temporal information, might enable better temporal integration.

    To investigate this, we finetune Qwen2.5-VL-7B on 5-second and 10-second videos using multiple temporal sampling rates, ranging from 1 fps to 10 fps, and evaluate cross-duration generalization. Each model is tested using the same fps setting used during training. The results are shown in Table~\ref{tab:fps_train}.

\begin{table}[t]
      \centering
      \small
      \begin{tabular}{c|cc|cc}
        \toprule
        & \multicolumn{2}{c|}{Train on 5s} & \multicolumn{2}{c}{Train on 10s} \\ 
        Train FPS & Test on 5s & Test on 10s & Test on 5s & Test on 10s \\ 
        \midrule 
        FPS-1 & 100.0 & 20.0 & 20.0 & 89.0 \\
        FPS-2 & 100.0 & 32.0 & 50.0 & 90.0 \\
        FPS-3 & 100.0 & 43.0 & 19.0 & 89.0 \\
        FPS-4 & 100.0 & 46.0 & 28.0 & 90.0 \\
        FPS-5 & 100.0 & 30.0 & 20.0 & 90.0 \\
        FPS-10 & 100.0 & 40.0 & 20.0 & 90.0 \\
        \bottomrule
      \end{tabular}
      \caption{Cross-generalization accuracy (\%) of fine-tuned models trained at various frame rates (FPS). Models are trained on either 5s or 10s videos and tested on both durations. For each evaluation, the test FPS is matched to the training FPS. The results show that increasing the training frame density does not solve the fundamental failure to generalize to unseen video length.}
      \label{tab:fps_train}
\end{table}

\begin{figure*}[t]
  \centering
  \includegraphics[width=0.95\linewidth]{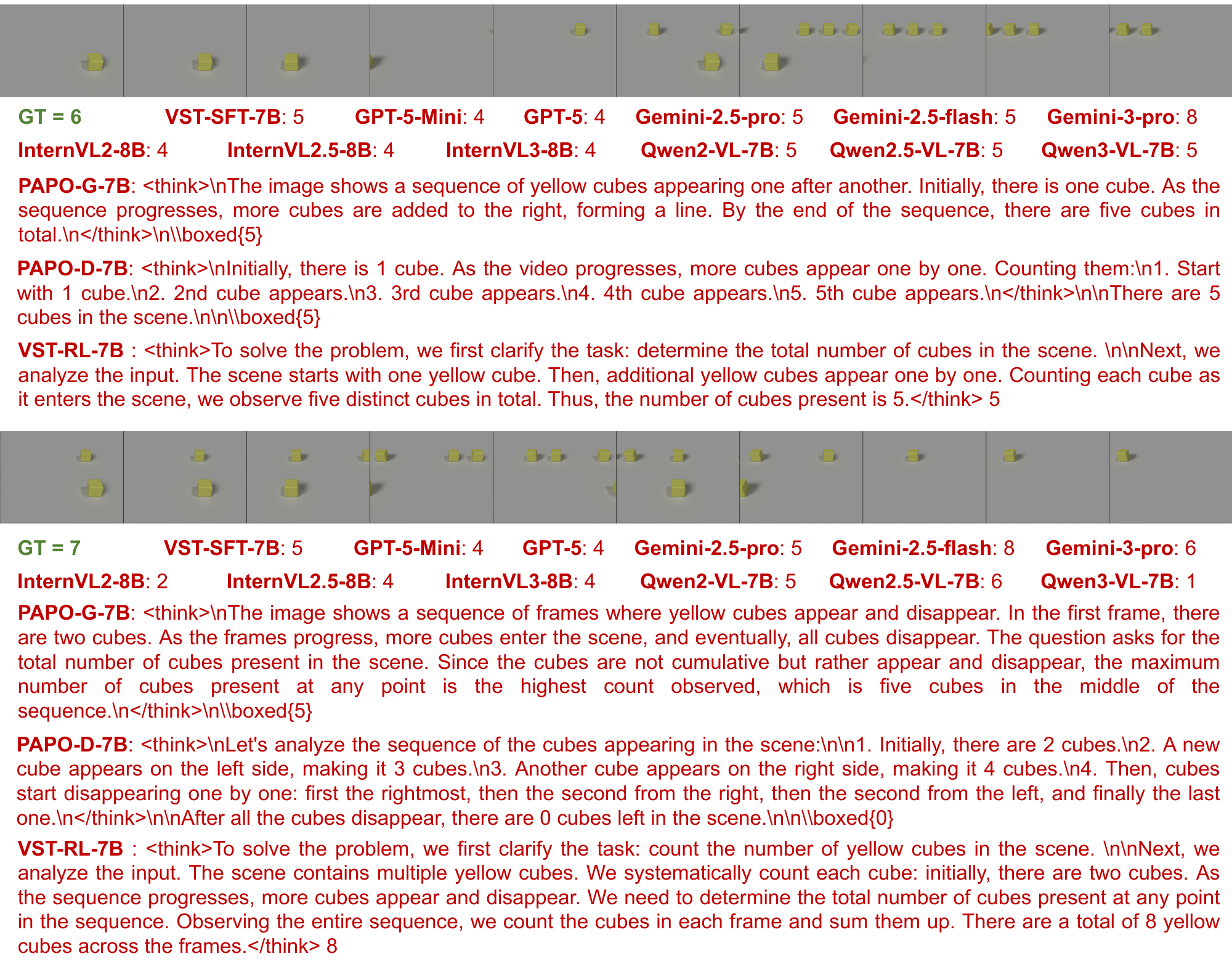}
  \caption{Two qualitative examples illustrating cases in which all evaluated models produce incorrect answers.}
  \label{fig:qualitative}
\end{figure*}

    Across all FPS settings, we observe a consistent pattern: training on 5-second videos yields perfect in-distribution performance (100\% accuracy) for all FPS values, but cross-generalization to 10-second test videos remains poor, achieving only 20–46\% depending on FPS; Training on 10-second videos also produces high in-distribution performance (89–90\%) but fails to generalize to 5-second test videos, with accuracies dropping to 19–50\%. These results mirror the earlier finetuning experiments, models can memorize or exploit shortcuts tied to the exact temporal structure of training data, but they fail to learn temporal aggregation mechanisms that are flexible across different video lengths.

    The failure of high-FPS training to improve generalization highlights a critical insight: \textbf{continuous perception is not simply a matter of providing more visual frames. It requires architectural or algorithmic mechanisms capable of building persistent, time-invariant representations}. Even when supplied with denser temporal information (up to 10 fps), the model does not develop the ability to track objects across arbitrarily long or short trajectories. This strongly suggests that the limitation lies not in the data but in the modeling paradigm itself—current architectures operate on fragmented frame-wise features rather than maintaining a unified, evolving scene representation.

\section{Conclusion}

    In this work, we argue that the prevailing paradigm of processing video as a collection of discrete, sampled frames is a fundamental barrier to achieving human-like visual intelligence. This approach leads to a fragmented and superficial understanding, in stark contrast to the holistic, continuous perception employed by humans. To empirically demonstrate this gap, we introduced the Continuous Perception Benchmark (CP-Bench), a minimalist diagnostic task where models must count visually identical objects in a panning video. This simple setup effectively neutralizes the texture-based shortcut solutions that many contemporary benchmarks permit, isolating the core capability of spatio-temporal correspondence. Despite the benchmark’s extreme simplicity, our experiments reveal that modern state-of-the-art multimodal models consistently fail on this trivially simple task. 

    We hope that this work opens a new research direction toward models capable of genuinely continuous perception, \ie systems that track persistent structure in the world, accumulate information over time, and reason about scenes the way humans do. Achieving this capability will likely require innovations in architecture, memory, temporal modeling, and training methodology. CP-Bench offers a stepping stone toward that goal, highlighting both the limitations of current approaches and the opportunities for progress in building physically grounded, temporally coherent multimodal intelligence.

{
    \small
    \bibliographystyle{ieeenat_fullname}
    \bibliography{main}
}


\end{document}